\documentclass[10pt,twocolumn,letterpaper]{article}

\usepackage[pagenumbers]{cvpr}
\usepackage{textgreek}
\usepackage{cuted}
\usepackage{xcolor}
\newcommand{\bluebf}[1]{\textcolor{blue}{\textbf{#1}}}

\newcommand{\todo}[1]{{\color{red}#1}}

\definecolor{cvprblue}{rgb}{0.21,0.49,0.74}
\usepackage[pagebackref,breaklinks,colorlinks,allcolors=cvprblue]{hyperref}

\def\confName{CVPR}
\def\confYear{2027}

\title{MaRSNav: Context-Aware Spatial Memory Retrieval and Revisable Task State for Continuous Vision-and-Language Navigation}

\author{
    Chenghao Li$^{1}$ \quad
    Shengyin Wang$^{1}$ \quad
    Mohan Ren$^{2}$ \quad
    Zhenghao Li$^{1}$ \quad
    Hujun Yin$^{1,*}$ \quad
    Xiaoxiao Cheng$^{1,*}$\\[4pt]
    $^1$The University of Manchester,
    $^2$The University of Queensland\\[2pt]
}

\begin{document}
\maketitle
\begingroup
\renewcommand{\thefootnote}{}
\footnotetext{
{\tt\small \{chenghao.li,shengyin.wang,hujun.yin, \\ xiaoxiao.cheng\}@manchester.ac.uk,
Zhenghao.li-4@postgrad.manchester.ac.uk,
uqmren2@uq.edu.au,
} \\ $^{*}$Corresponding authors. }
\endgroup



\begin{abstract}

Continuous vision-and-language navigation (VLN) requires agents to leverage historical spatial information while accurately tracking their progress through multi-step instructions. However, directly incorporating historical observations into vision-language models (VLMs) increases input sequence length and self-attention cost, while errors in task-progress estimation can propagate across navigation decisions. 
To address these challenges, we propose \textbf{MaRSNav}, a unified framework that integrates task-conditioned spatial memory retrieval, revisable subtask tracking, and grounded navigation prediction. 
A frozen geometry encoder constructs external spatial memory from multi-view history, which the VLM retrieves through context-conditioned decoder-level cross-attention.
A persistent subtask state tracks navigation progress, with training-time perturbations encouraging correction of inaccurate estimates. 
Role-specific queries jointly predict image-space waypoints and local trajectories in a single forward pass, supported by auxiliary future-feature supervision.
Trained solely on R2R-CE and RxR-CE navigation demonstrations, without DAgger or additional navigation training data, MaRSNav achieves success rates of \textbf{65.4\%} and \textbf{66.5\%} on their respective val-unseen splits, outperforming several methods trained with over 10 million additional samples.

\end{abstract}    
\section{Introduction}
\label{sec:introduction}

Vision-and-language navigation (VLN) requires an embodied agent to follow a natural-language instruction through a previously unseen environment using only its visual observations ~\cite{anderson2018r2r,krantz2020vlnce, ku2020rxr}. Early VLN approaches developed task-specific policies for instruction following and navigation history encoding~\cite{fried2018speaker,hong2021recurrentvlnbert,chen2021hamt}. Recently, these tasks have advanced rapidly since pretrained vision-language models (VLMs) were adopted as navigation policies. The VLMs now enable agents to leverage rich semantic knowledge to act on free-form natural-language instructions directly from egocentric visual observations~\cite{zhang2024navid,zhang2025uninavid,cheng2025navila,wei2026dualvln}.

Despite these advances, significant gap persists in long-horizon tasks where navigation decisions may depend on landmarks and spatial relationships encountered earlier but no longer visible in the current observations.
Existing approaches preserve historical information through compressed visual tokens~\cite{zhang2024navid,zhang2025uninavid} or geometry-aware visual representations~\cite{wang2025dynam3d,zeng2026janusvln}.
Incorporating historical observations into the VLM's input sequence increases its length and the associated self-attention cost. In addition, the relevance of historical information depends on navigation instruction, current observations, and progress of the task. These considerations motivate an external memory mechanism that enables context-dependent retrieval of historical spatial information without expanding the VLM's input sequence with past observations.


While spatial memory preserves information about previously observed regions, following a multi-stage instruction also requires tracking which subtasks have been completed. Existing studies generally fall into two categories. One approaches infer task progress from observation history or estimate it through an auxiliary prediction objective~\cite{ma2019selfmonitoring}. The other introduces  Chain-of-Thought (CoT)\cite{xue2026omninav, huang2026ticvla, zhao2025cotvla}, forcing the model to reason and generalize based on information to determine task progress. An explicit task state can maintain a record of completed and remaining subtasks across navigation decisions. However, as this state relies on the model's own predictions, incorrect completion estimates may persist and affect subsequent decisions. Effective task-state tracking therefore requires not only maintaining progress information but also revising previous estimates when new visual evidence indicates an error.

To address these challenges, we propose \textbf{MaRSNav}, a continuous vision-and-language navigation framework that integrates external spatial memory with persistent and revisable task-state tracking.
A frozen geometry encoder extracts spatial features from recent multi-view observations~\cite{wang2026vggtomega}, which the VLM accesses through decoder-level cross-attention~\cite{alayrac2022flamingo}, conditioned on the navigation instruction, current observations, and previously maintained task state.
This design enables the VLM to retrieve historical spatial information without incorporating past image tokens into its input sequence.
Meanwhile, the VLM maintains an ordered subtask plan, updating the completion status at each decision and carrying the revised state forward.
To prevent inaccurate progress estimates from propagating, we perturb the input task state during training while supervising the correct state, encouraging the model to revise earlier predictions based on visual evidence.

The resulting spatial and task context also guides navigation planning, where specialised prediction heads jointly generate an image-space waypoint~\cite{wei2026dualvln} and a local path.
An auxiliary future-feature prediction objective, supervised by a frozen V-JEPA~2 encoder~\cite{assran2025vjepa2}, further supports representation learning during training.
During execution, a controller follows the predicted path, while the VLM periodically replans using updated observations and task state.
Without additional navigation training data beyond R2R-CE and RxR-CE demonstrations~\cite{krantz2020vlnce,ku2020rxr}, MaRSNav achieves success rates of $65.4\%$ and $66.5\%$ on their respective val-unseen splits.

\noindent In summary, the main contributions of this paper include:
\begin{itemize}
    \item A context-aware spatial memory retrieval scheme that constructs external spatial memory from multi-view history using a frozen geometry encoder and enables the VLM to retrieve relevant historical spatial information through decoder-level cross-attention, without adding historical image tokens to its input sequence.
 
    \item A task progress tracking mechanism that maintains a record of completed and remaining subtasks across navigation decisions, with state perturbation during training to encourage correction of inaccurate progress estimates using visual evidence.
 
    \item A joint prediction and supervision design that integrates task-progress updating, waypoint localisation, and local trajectory prediction within a unified VLM architecture, supported by auxiliary future-latent supervision.

\end{itemize}

\section{Related Work}
\label{sec:related}

\paragraph{Spatial memory representation and retrieval.}
VLM-based agents usually keep history in the token sequence: video-based
models compress each past frame into a few visual
tokens~\cite{zhang2024navid,zhang2025uninavid}, so sequence length and
self-attention cost grow with the history.
Geometry-aware models add 3D structure to these tokens, from depth or 3D
inputs~\cite{hong2023_3dllm,zheng2025video3dllm,zhu2025llava3d,wang2025dynam3d}
or from feed-forward geometry models such as
VGGT~\cite{wang2025vggt,wang2026vggtomega}.
How the geometry reaches the language model differs: Spatial-MLLM and
JanusVLN fuse it with the semantic tokens before the language
model~\cite{wu2025spatialmllm,zeng2026janusvln}, and SpatialStack adds it
at the visual-token positions inside the
decoder~\cite{zhang2026spatialstack}.
In both cases geometry is attached to tokens already in the sequence, so
the language model receives fused features rather than a memory it can
query.
MaRSNav keeps the multi-view history as an external spatial memory
that every decoder state queries by
cross-attention~\cite{alayrac2022flamingo}. Retrieval thus depends on
the current context, and past views are reached without entering the
sequence.

\paragraph{Task progress tracking and state revision.}
Following a multi-stage instruction requires knowing which parts are
already done.
Early agents estimated progress as an auxiliary
signal~\cite{ma2019selfmonitoring}. VLM-based agents express it in
language, through learned reasoning
steps~\cite{lin2025navcot,wang2025auxthink}, subtask selection by a slow
system~\cite{xue2026omninav}, or by writing out completed and upcoming
sub-instructions~\cite{liu2026navforesee}.
A progress state that persists across decisions raises a further
problem: a wrong completion estimate is fed back and can mislead later
decisions.
Error correction has mostly targeted execution: CorrectNav learns to
recover from deviations by collecting error
trajectories~\cite{yu2026correctnav}.
MaRSNav carries the whole ordered plan with its completion marks from
one decision to the next and perturbs this state on expert samples
during training, so the model learns to revise it from visual evidence
without additional rollouts.

\paragraph{From decision to motion.}
Most VLM-based agents output low-level actions or mid-level commands as
text~\cite{zhang2024navid,zhang2025uninavid,cheng2025navila}.
Dual-system methods instead hand a goal to a separate generative policy,
a diffusion or flow-matching model conditioned on a grounded pixel, a
selected subtask, or geometry
queries~\cite{wei2026dualvln,xue2026omninav,peng2026logoplanner}.
Predicted futures serve as additional supervision, in the form of
subgoal images~\cite{zhao2025cotvla}, future depth and semantic
features~\cite{liu2026navforesee}, or visual targets for role-specific
queries~\cite{zhang2025dreamvla}.
MaRSNav takes its image-space waypoint from DualVLN and its
query-based future supervision from DreamVLA, but obtains goal and
motion from one model: in a single forward pass, role queries yield the
waypoint and a directly regressed local trajectory, without a second
policy or iterative sampling.

\section{Methodology}
\label{sec:method}
 
\begin{figure*}[t]
    \centering
    \includegraphics[
        width=1.0\textwidth,
        clip
    ]{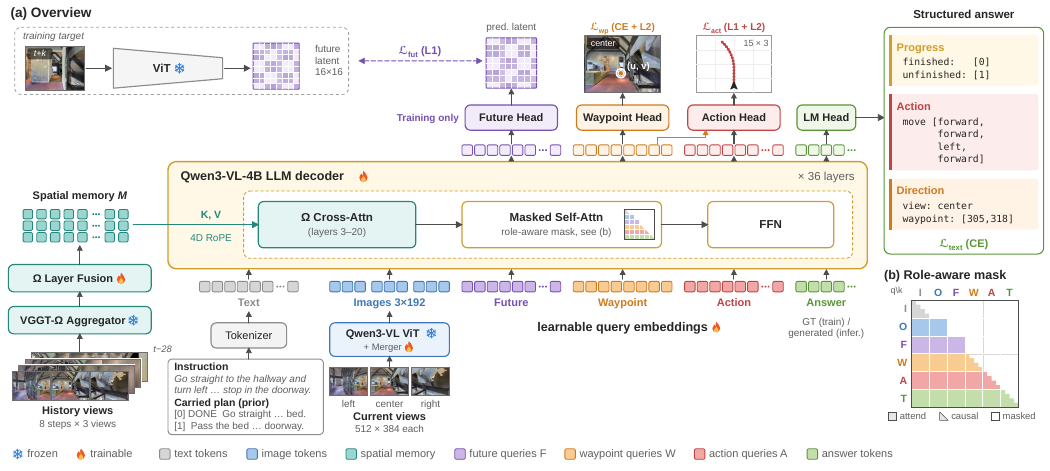}
    \caption{\textbf{MaRSNav.}
    \textbf{(a)} Historical multi-view observations form an external
    spatial memory, while the VLM processes the instruction, the subtask
    state, and the current views. Injected decoder layers read the memory
    before self-attention. 4D RoPE rotates the memory keys only.
    Future ($F$), waypoint ($W$), and action ($A$) queries serve the
    auxiliary future-feature loss, waypoint grounding, and local
    trajectory regression. The waypoint and trajectory are read out in
    one forward pass over the prompt. The structured answer is then
    decoded autoregressively. The Future Head and the frozen V-JEPA~2
    encoder (ViT) are used only during training.
    $\mathcal L_{\rm wp}$ sums a view classification loss (CE) and a
    squared pixel error (L2). In $\mathcal L_{\rm act}$, L1 supervises
    normalized increments and L2 denotes unsquared path and endpoint
    distances.
    \textbf{(b)} The role-aware mask controls information flow among
    the input context, the role queries, and the answer tokens.}
    \label{fig:tbground}
\end{figure*}
 
\subsection{Overview}
\label{sec:method_overview}
 
MaRSNav pairs a VLM that reasons over the current views with a frozen
VGGT \cite{wang2026vggtomega}geometry model that remembers the past. The overview model architecture is shown in \cref{fig:tbground}(a).
At a decision made at environment step $t$, it receives the instruction
$\mathcal I$, the current left, center, and right RGB views $O_t$, the
subtask state $s_t$, and a sampled observation history $\mathcal H_t$.
Qwen3-VL-4B~\cite{bai2025qwen3vl} processes $\mathcal I$, $s_t$, and
$O_t$, while $\mathcal H_t$ enters only through the spatial memory
(\cref{sec:geometric_memory}).
Learnable role queries appended to the prompt are read out as an
image-space waypoint and a local trajectory in a single forward pass
over the prompt (\cref{sec:role_queries}). The language head then
decodes a structured answer that updates
the subtask state from the same key-value cache, which is shown in \cref{sec:subtask_state}.
 
\subsection{Spatial Memory}
\label{sec:geometric_memory}
\paragraph{Construction.}
We sample eight timepoints at a stride of four environment steps, each
with left, center, and right views and including the current
observation. Missing early observations repeat the initial frame.
A frozen VGGT-$\Omega$~\cite{wang2026vggtomega} encodes these views
jointly.
From layers $
\mathcal R=\{4,11,17,23\}$ we take its features
$G_t^{(r)}$, drop the camera tokens, and keep the register and patch
tokens of every view.
The $\Omega$ Layer Fusion normalizes and projects each layer
independently and concatenates them along channels:
\begin{equation}
M_t=\operatorname{Flatten}_{n,p}\!\left[
\mathop{\operatorname{Concat}}_{r\in\mathcal R}
P_r\!\left(\operatorname{LN}_r(G_t^{(r)})\right)
\right]
\label{eq:memory_fusion}
\end{equation}
where $n$ indexes views, $p$ indexes tokens, and $P_r$ is a learned
linear projection.
\vspace{-1em}
\paragraph{View- and time-aware keys.}
Each decoder layer $\ell$ that reads the memory normalizes the $i$-th
memory token, $m_i^\ell=\operatorname{LN}_M^\ell(M_{t,i})$, and projects
it with learned matrices $W_K^\ell$ and $W_V^\ell$ to a key and a value:
\begin{equation}
K_i^\ell=\operatorname{RoPE}_{\rho_i}(m_i^\ell W_K^\ell),
\qquad
V_i^\ell=m_i^\ell W_V^\ell.
\label{eq:memory_rope}
\end{equation}
Here $\rho_i=(v_i,\tau_i,h_i,w_i)$ is the view, timepoint, patch row,
and patch column of token $i$ ($\rho_i=\mathbf0$ for VGGT's register tokens),
and $\operatorname{RoPE}_{\rho_i}$ is a four-axis rotary position
encoding (4D RoPE) that rotates pairs of key channels by angles
proportional to these coordinates, one group of channels per axis.
Queries and values are not rotated, as the querying tokens have no
position in this grid.
 
We insert cross-attention to $(K^\ell,V^\ell)$ in decoder layers 3--20
of 36 (zero-based), after Qwen3-VL's initial DeepStack visual
integrations~\cite{bai2025qwen3vl}.
Each injected layer first reads the memory and then applies
self-attention (both choices are ablated in our experiments), so retrieved evidence is exchanged among tokens within
the layer and conditions the retrieval of the next.
Every position $j$, whether text, image, or role query, queries the
memory with its own normalized hidden state $\bar h_j^\ell$:
\begin{equation}
h_j^\ell \leftarrow h_j^\ell +
\operatorname{Attn}\bigl(\bar h_j^\ell W_Q^\ell,\,K^\ell,\,V^\ell\bigr)W_O^\ell ,
\label{eq:memory_read}
\end{equation}
where $\operatorname{Attn}$ is scaled dot-product attention and
$W_O^\ell$ is initialized to zero, so the residual is inactive at the
start of training without a gate.
The query has no position of its own. What it retrieves depends on
$h_j^\ell$, which by layer~3 has already attended to the tokens before
it: the instruction and the subtask state for image tokens, and
additionally the current views for role queries.
Retrieval is context-aware in this sense.
 
\subsection{Subtask State}
\label{sec:subtask_state}\paragraph{State and answer format.}
The subtask state $s_t$ is a short text: an ordered plan of subtasks,
each marked finished or unfinished.
At every decision the model receives $s_t$ and outputs a structured
answer with three fields:
\emph{progress} (the indices of finished and unfinished subtasks),
\emph{action} (a movement sequence, look-down, or stop), and
\emph{direction} (the target view and pixel coordinates, when available).
The first answer additionally writes the plan, and the plan with the
updated progress is carried to the next decision.
The answer is trained with next-token cross-entropy
$\mathcal L_{\rm text}$ on answer tokens only.
 
All supervision is derived from the R2R-CE and RxR-CE training episodes.
No additional trajectories are collected.
Qwen3.7\textbf{ Plus and Max} splits each
instruction into an ordered list of subtasks and marks the frame of the
expert trajectory at which each subtask is completed. We treat this
boundary as a window of $\pm3$ frames.
At step $t$, the subtasks completed before $t$ are marked finished,
which gives $s_t$ and the progress target.
The waypoint target is the expert-path position
projected into the current views with
the simulator's camera model, and its view label is the view in which it
falls.
The label is \emph{no-goal} when the projection lies outside all views
or during a turning maneuver such as turning around, and \emph{down}
follows the look-down rule of DualVLN~\cite{wei2026dualvln}.
The policy itself never receives depth or camera parameters.
 
During training, the carried state is the ground-truth state of the
previous decision. At inference it is the model's own earlier output, so
an error in it would be fed back unchanged.
We therefore draw each training sample under one of four prompt
variants.
One asks the model to write the plan itself, as it must at the first
decision.
The other three provide the plan with a clean state, a lightly perturbed
state, or a heavily perturbed state
, while the observations, the action targets,
and the target state stay unchanged.
Under perturbation the model must recover the correct state from the
instruction, the current views, and the spatial memory, which makes its
progress output depend on what it observes rather than copy what it
carries.
 
\subsection{Role Queries and Readout}
\label{sec:role_queries}
 
Learnable future ($F$), waypoint ($W$), and action ($A$) queries are
placed after the prompt and before the answer tokens ($T$).
The attention mask is shown in \cref{fig:tbground}(b), where $I$ is the
text prefix and $O$ the current observation context (image tokens and
interleaved view text). It is the same in training and inference.
Role queries cannot read answer tokens, so teacher-forced answers never
leak into them.
Future queries read the context and one another but not $W$ or $A$.
Waypoint queries additionally read $F$, and action queries read $F$ and
$W$, with causal attention inside the $W$ and $A$ blocks.
Like all positions, role queries also read the spatial memory directly.
 
Following DualVLN's pixel-goal grounding~\cite{wei2026dualvln}, the local
target is an image-space waypoint.
An attention-pooling readout over the normalized final waypoint states
$H_t^W$, followed by an MLP, predicts view logits $\hat y_t$ over
\{left, center, right, down, no-goal, stop\} and normalized pixel
coordinates $\hat u_t$.

\begin{equation}
\mathcal L_{\rm wp}
=\underbrace{\operatorname{CE}(\hat y_t,y_t^*)}_{\mathcal L_{\rm view}}
+\underbrace{\operatorname{MSE}_{\rm valid}(\hat u_t,u_t^*)}_{\mathcal L_{\rm uv}},
\label{eq:waypoint_loss}
\end{equation} 
where no-goal and stop targets are excluded from $\mathcal L_{\rm uv}$.
As in DualVLN \cite{wei2026dualvln}, when the answer requests a look-down, a downward view
replaces the center image and waypoint grounding is rerun. The spatial
memory keeps its horizontal views.
 
The Action Head is a three-layer transformer decoder that attends to the
concatenated states $[H_t^W;H_t^A]$ and regresses, in one pass, $N=15$
increments $a_i=(\Delta x_i,\Delta y_i,\Delta\psi_i)$ in the agent frame
at prediction time ($x$ forward, $y$ left).
Targets follow the expert path toward the waypoint, resampled at about
$0.2$\,m up to $3$\,m and padded with zero increments after arrival.
Because the waypoint already fixes the local goal, the remaining motion
is short-range, and we regress it deterministically without iterative
sampling.
With $\tilde a_i$ the per-component normalized increments and
$\hat p_i=\sum_{j\le i}\hat a_{j,xy}$ the accumulated positions,
\begin{equation}
\begin{aligned}
\mathcal L_\Delta
    &= \tfrac{1}{3N}\textstyle\sum_{i=1}^{N}
    \|\hat{\tilde a}_{i}-\tilde a^*_{i}\|_1, \\
\mathcal L_{\rm path}
    &= \tfrac{1}{N}\textstyle\sum_{i=1}^{N}
    \|\hat p_{i}-p^*_{i}\|_2,\\
\mathcal L_{\rm end}
    &= \|\hat p_{N}-p^*_{N}\|_2,\\
\mathcal L_{\rm act}
    &= 2\mathcal L_\Delta+2\mathcal L_{\rm path}
    +0.5\mathcal L_{\rm end}
\end{aligned}
\label{eq:trajectory_losses}
\end{equation}
averaged over the batch. The path and endpoint terms are unsquared
distances in meters.
These gradients reach the waypoint states, the decoder, and the spatial
memory cross-attention.
 
\paragraph{Auxiliary future-latent loss.}
The waypoint and trajectory targets supervise only a few numbers per
decision.
To give the memory read a denser signal, the future queries predict the
features of an upcoming view: what the agent will see after it advances
depends on the layout ahead more than on the appearance of the current
views, and the spatial memory is where layout is encoded explicitly.
A shared MLP maps the normalized states $H_t^F$ to a $16{\times}16$
feature grid $\hat Z_t$, and a frozen V-JEPA~2 encoder
$E$~\cite{assran2025vjepa2} provides the target from the center view
$o^{\rm c}$, $k$ steps ahead:
\begin{equation}
\mathcal L_{\rm fut}
    =\operatorname{MAE}_{\rm valid}\!\bigl(\hat Z_t,\,
    \operatorname{sg}[E(o^{\rm c}_{t+k})]\bigr)
\label{eq:future_loss}
\end{equation}
where $\operatorname{sg}$ stops gradients.
The offset $k$ is the frame associated with the waypoint when available
and three steps otherwise. Stop samples use the current frame, and
missing targets are masked.
The loss acts in latent space and is purely auxiliary: nothing is
decoded to pixels, the Future Head is discarded at inference, and only
the future queries remain, whose states the waypoint and action queries
read through the role-aware mask.
 
\subsection{Training and Execution}
\label{sec:heads_training}
 
The full objective is
\begin{equation}
\mathcal L=1.5\,\mathcal L_{\rm text}
    +2\,\mathcal L_{\rm wp}
    +0.5\,\mathcal L_{\rm fut}
    +\mathcal L_{\rm act},
\label{eq:total_loss}
\end{equation}
with each term averaged over its valid targets.
Training has two stages: the first omits the action branch and uses
$\mathcal L_{\rm text}$, $\mathcal L_{\rm wp}$, and
$\mathcal L_{\rm fut}$. The second adds the action queries and the
Action Head and optimizes all terms.
VGGT-$\Omega$, the VLM vision backbone, and the V-JEPA~2 encoder stay
frozen. The decoder, the visual merger, and the added modules are
trained. In inference, a controller follows the predicted XY path using measured poses, the history is updated after each action, and the model is
queried again after at most three environment actions.
Structured outputs handle the initial discrete movement, the no-goal
fallback, look-down refinement, and stopping.

\providecommand{\method}{Ours}
\newcommand{\ck}{\checkmark}
\newcommand{\nr}{n/r}

\begin{table*}[]
\centering
\caption{\textbf{Comparison on VLN-CE val-unseen.}
\emph{Pano.}: panoramic or multi-camera input. \emph{S.RGB}: a single RGB camera.
\emph{External data}: training samples beyond the R2R-CE and RxR-CE training splits
(\eg, ScaleVLN, DAgger rollouts, question answering), listed for RGB-only methods
following~\cite{zeng2026janusvln,zhang2026futurenav}. \nr: not reported.
Best results are in \bluebf{blue bold}, second best in \textbf{bold}.}
\label{tab:main}
\footnotesize
\setlength{\tabcolsep}{3pt}
\resizebox{\textwidth}{!}{%
\begin{tabular}{l cccc cccc ccc r}
\toprule
& \multicolumn{4}{c}{Observation}
& \multicolumn{4}{c}{R2R-CE Val-Unseen}
& \multicolumn{3}{c}{RxR-CE Val-Unseen}
& External \\
\cmidrule(lr){2-5}\cmidrule(lr){6-9}\cmidrule(lr){10-12}
Method & Pano. & Odo. & Depth & S.RGB
& NE$\downarrow$ & OSR$\uparrow$ & SR$\uparrow$ & SPL$\uparrow$
& NE$\downarrow$ & SR$\uparrow$ & SPL$\uparrow$
& data \\
\midrule
HPN+DN~\cite{krantz2021hpn}      & \ck & \ck & \ck &     & 6.31 & 40.0 & 36.0 & 34.0 & --    & --   & --   & -- \\
CMA~\cite{hong2022bridging}           & \ck & \ck & \ck &     & 6.20 & 52.0 & 41.0 & 36.0 & 8.76  & 26.5 & 22.1 & -- \\
VLN$\circlearrowright$BERT~\cite{hong2022bridging} & \ck & \ck & \ck & & 5.74 & 53.0 & 44.0 & 39.0 & 8.98 & 27.0 & 22.6 & -- \\
Sim2Sim~\cite{krantz2022sim2sim}      & \ck & \ck & \ck &     & 6.07 & 52.0 & 43.0 & 36.0 & --    & --   & --   & -- \\
GridMM~\cite{wang2023gridmm}          & \ck & \ck & \ck &     & 5.11 & 61.0 & 49.0 & 41.0 & --    & --   & --   & -- \\
DreamWalker~\cite{wang2023dreamwalker}& \ck & \ck & \ck &     & 5.53 & 59.0 & 49.0 & 44.0 & --    & --   & --   & -- \\
Reborn~\cite{wang2022reborn}            & \ck & \ck & \ck &     & 5.40 & 57.0 & 50.0 & 46.0 & 5.98  & 48.6 & 42.0 & -- \\
ETPNav~\cite{an2025etpnav}            & \ck & \ck & \ck &     & 4.71 & 65.0 & 57.0 & 49.0 & 5.64  & 54.7 & 44.8 & -- \\
ScaleVLN~\cite{wang2023scalevln}      & \ck & \ck & \ck &     & 4.80 & --   & 55.0 & 51.0 & --    & --   & --   & -- \\
HNR~\cite{wang2024hnr}                & \ck & \ck & \ck &     & 4.42 & 67.0 & 61.0 & 51.0 & 5.50  & 56.3 & 46.7 & -- \\
\midrule
AG-CMTP~\cite{chen2021cmtp}                 & \ck & \ck & \ck &     & 7.90 & 39.0 & 23.0 & 19.0 & --    & --   & --   & -- \\
R2R-CMTP~\cite{chen2021cmtp}                & \ck & \ck & \ck &     & 7.90 & 38.0 & 26.4 & 22.7 & --    & --   & --   & -- \\
InstructNav~\cite{long2024instructnav}      & \ck & \ck & \ck &     & 6.89 & --   & 31.0 & 24.0 & --    & --   & --   & -- \\
AO-Planner~\cite{chen2025aoplanner}         & \ck &     & \ck &     & 5.55 & 59.0 & 47.0 & 33.0 & 7.06  & 43.3 & 30.5 & -- \\
\midrule
Seq2Seq~\cite{krantz2020vlnce}              &     &     & \ck & \ck & 7.77 & 37.0 & 25.0 & 22.0 & 12.10 & 13.9 & 11.9 & -- \\
CMA~\cite{krantz2020vlnce}                  &     &     & \ck & \ck & 7.37 & 40.0 & 32.0 & 30.0 & --    & --   & --   & -- \\
LAW~\cite{raychaudhuri2021law}              &     & \ck & \ck & \ck & 6.83 & 44.0 & 35.0 & 31.0 & 10.90 & 8.0  & 8.0  & -- \\
CM2~\cite{georgakis2022cm2}                 &     & \ck & \ck & \ck & 7.02 & 41.5 & 34.3 & 27.6 & --    & --   & --   & -- \\
WS-MGMap~\cite{chen2022wsmgmap}             &     & \ck & \ck & \ck & 6.28 & 47.6 & 38.9 & 34.3 & --    & --   & --   & -- \\
Dynam3D~\cite{wang2025dynam3d}              &     & \ck & \ck & \ck & 5.34 & 62.1 & 52.9 & 45.7 & --    & --   & --   & -- \\
D3D-VLP~\cite{wang2026d3dvlp}               &     & \ck & \ck & \ck & 4.73 & 67.2 & 61.3 & 56.1 & --    & --   & --   & -- \\
\midrule
NaVid~\cite{zhang2024navid} & & & & \ck & 5.47 & 49.1 & 37.4 & 35.9 & --   & --   & --   & 953K \\ 
Uni-NaVid~\cite{zhang2025uninavid} & & & & \ck & 5.58 & 53.3 & 47.0 & 42.7 & 6.24 & 48.7 & 40.9 & 3577K \\ 
NaVILA~\cite{cheng2025navila} & & & & \ck & 5.22 & 62.5 & 54.0 & 49.0 & 6.77 & 49.3 & 44.0 & 13132K \\ 
DecoVLN~\cite{xin2026decovln} & & & & \ck & 5.01 & 63.5 & 56.3 & 50.5 & 5.73 & 54.2 & 46.3 & $>$180K \\ 
StreamVLN~\cite{wei2026streamvln} & & & & \ck & 4.98 & 64.2 & 56.9 & 51.9 & 6.22 & 52.9 & 46.0 & $\sim$26330K \\ 
NavFoM (single view)~\cite{zhang2026navfom} & & & & \ck & 5.01 & 64.9 & 56.2 & 51.2 & 5.51 & 57.4 & 49.4 & $\sim$10480K \\ 
JanusVLN (0K)~\cite{zeng2026janusvln} & & & & \ck & 5.17 & 58.0 & 52.8 & 49.2 & 6.46 & 51.4 & 44.3 & 0K \\ 
JanusVLN~\cite{zeng2026janusvln} & & & & \ck & 4.78 & 65.2 & 60.5 & 56.8 & 6.06 & 56.2 & 47.5 & 10692K \\ 
DualVLN~\cite{wei2026dualvln} & & & & \ck & \textbf{4.05} & 70.7 & 64.3 & 58.5 & 4.58 & 61.4 & 51.8 & $\sim$11000K  \\ 
CorrectNav~\cite{yu2026correctnav} & & & & \ck & 4.24 & 67.5 & 65.1 & \textbf{62.3} & \textbf{4.09} & \textbf{69.3} & \bluebf{63.3} & $>$240K \\ 
\midrule
NavFoM ~\cite{zhang2026navfom} & \ck & & & & 4.61 & \textbf{72.1} & 61.7 & 55.3 & 4.74 & 64.4 & 56.2 & $\sim$10480K \\ 
OmniNav~\cite{xue2026omninav} & \ck & & & & \bluebf{3.74} & \bluebf{74.6} & \bluebf{69.5} & \bluebf{66.1} & \bluebf{3.77} & \bluebf{73.6} & \textbf{62.0} & $\sim$11400K \\ 
\midrule
\textbf{\method{}} & \ck & & & & 4.35 & 71.4 & \textbf{65.4} & 60.7 & 4.69   & 66.5   & 55.7   & 0K \\
\bottomrule
\end{tabular}%
}
\end{table*}


\section{Experiments}
\label{sec:experiment}

\subsection{Experimental Setup}
\label{sec:exp_setup}\paragraph{Benchmarks and metrics.}
We evaluate MaRSNav on R2R-CE and RxR-CE in continuous Matterport3D
environments using Habitat~\cite{anderson2018r2r,krantz2020vlnce,ku2020rxr}.
R2R-CE contains 10,819 training and 1,839 val-unseen episodes. The
RxR-CE English guide subset (en-US and en-IN) contains 19,996 training
and 3,669 val-unseen episodes.
On val-unseen we report navigation error (NE), success rate (SR, a stop
within $3$\,m of the goal), oracle success rate (OSR), and success
weighted by path length (SPL).

\vspace{-1em}
\paragraph{Environment setup.}
The agent observes left, center, and right RGB views ($\pm60^\circ$ yaw,
$90^\circ$ FOV) from a camera height of $0.85$\,m, plus a downward view
for look-down refinement.
Views are rendered at $640\times480$ and resized to $512\times384$
before entering the model. Pixel coordinates are mapped in the
rendered frame.
Actions comprise $0.25$\,m forward steps, $15^\circ$ turns, and STOP.
The model is queried again after at most three executed actions, and
episodes are capped at $400$ steps.
The controller tracks predicted trajectories with simulator pose
feedback. The policy itself receives no pose, depth, or map.
\vspace{-1em}

\paragraph{Implementation details.}
MaRSNav builds on Qwen3-VL-4B~\cite{bai2025qwen3vl} and
VGGT-$\Omega$~\cite{wang2026vggtomega} and follows the two training
stages of \cref{sec:heads_training}, without DAgger or reinforcement
learning.
The joint-training set contains $1.63$M samples from the R2R-CE and
RxR-CE training data.
We use AdamW on four H200 GPUs with an effective batch size of $144$ and
peak learning rates of $10^{-5}$ for the language decoder, $10^{-4}$ for
the spatial memory cross-attention and the prediction heads, and
$10^{-6}$ for the visual merger.
The reported checkpoint is taken after $8$K updates of the second stage
($0.71$ epochs, about $56$ hours).

%
\providecommand{\todo}[1]{\textcolor{red}{[#1]}}  

\subsection{Main Results}
\label{sec:main_results}

\paragraph{Comparison with prior methods.}
Table~\ref{tab:main} reports val-unseen results together with the
external training data of each RGB-only method.
Trained only on R2R-CE and RxR-CE demonstrations, MaRSNav reaches
$65.4\%$ SR and $60.7\%$ SPL on R2R-CE, and $66.5\%$ SR and $55.7\%$
SPL on RxR-CE.
Every other RGB-only entry in Table~\ref{tab:main} above $60\%$ SR on
R2R-CE uses external data, from $240$K samples
(CorrectNav~\cite{yu2026correctnav}) to more than $10$M
(JanusVLN~\cite{zeng2026janusvln}, NavFoM~\cite{zhang2026navfom},
OmniNav~\cite{xue2026omninav}).
Without such data, MaRSNav is on par with CorrectNav in R2R-CE SR
($65.4$ \vs $65.1$), exceeds DualVLN by $1.1$ SR and $2.2$ SPL points on
R2R-CE and by $5.1$ and $3.9$ points on RxR-CE, and exceeds the full
JanusVLN model, trained with $10.7$M external samples, by $4.9$ SR
points on R2R-CE and $10.3$ on RxR-CE.
\vspace{-1em}
\paragraph{Benchmark-only training.}
Under the same data constraint, MaRSNav improves over
JanusVLN~(0K) by $12.6$ SR and $11.5$ SPL points on R2R-CE and by
$15.1$ and $11.4$ points on RxR-CE.
JanusVLN uses a single view and MaRSNav three, so we also compare
within the multi-view group: MaRSNav exceeds the four-view NavFoM by
$3.7$ SR and $5.4$ SPL points on R2R-CE and by $2.1$ SR points on
RxR-CE, where its SPL is $0.5$ points lower, against roughly $10.5$M
external samples for NavFoM.
OmniNav, which shares our three-view setup, remains ahead on both
benchmarks with about $11.4$M external samples from navigation,
question-answering, and grounding datasets, and CorrectNav is ahead on
RxR-CE with self-correction rollouts.
A policy trained only on demonstrations never observes its own
deviations, and we regard such data as complementary to the architecture
studied here.



\begin{table}[t]
\centering
\caption{\textbf{History length and number of role queries} on R2R-CE
val-unseen with models trained on R2R-CE only \textbf{(without sub-tasks state)}. The memory is read
before self-attention in layers $3$--$20$.
\emph{Frames}: history timepoints in the spatial memory.
\emph{Queries}: number of role queries.}
\label{tab:ablation}
\footnotesize
\setlength{\tabcolsep}{5pt}
\begin{tabular}{c c cccc}
\toprule
Frames & Queries & NE$\downarrow$ & OSR$\uparrow$ & SR$\uparrow$ & SPL$\uparrow$ \\
\midrule
6 & $N$ & 7.23 & 51.82 & 40.29 & 36.73 \\ 
8 & $N$ & \textbf{7.20} & \textbf{53.18} & \textbf{42.25} & \textbf{38.37} \\ 
9 & $N$ & 7.45 & 52.20 & 40.73 & 36.89 \\ 
\midrule
4 & $N/4$ & 7.58 & 49.76 & 36.76 & 33.18 \\ 
8 & $N/4$ & 7.57 & 50.63 & 40.29 & 36.78 \\ 
\bottomrule
\end{tabular}
\end{table}

\begin{figure}[t]
\centering
\includegraphics[width=\linewidth]{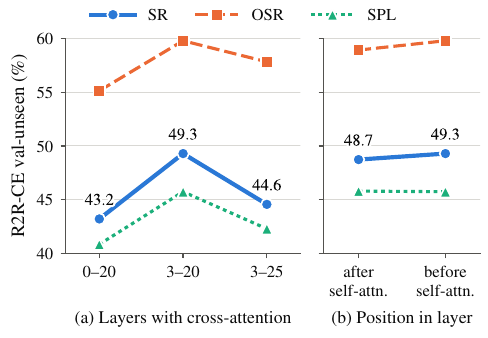}
\caption{\textbf{Spatial memory cross-attention ablations} on R2R-CE
val-unseen. Each model is trained on R2R-CE \textbf{(with sub-tasks state)} only.
(a)~Decoder layers that read the memory ($36$ in total).
(b)~Position within a layer, for layers $3$--$20$.}
\label{fig:ablation_layers}
\end{figure}

\subsection{Ablation Studies}
\label{sec:ablation}
All ablation models are trained on R2R-CE demonstrations
and evaluated on all $1{,}839$ R2R-CE val-unseen
episodes.
The agent moves towards the predicted waypoint for up to
four to eight actions before the model is queried again.
This setting is used only here and identically for every variant.
Absolute numbers are therefore not comparable with
Table~\ref{tab:main}.
The models in Fig.~\ref{fig:ablation_layers} are trained with the
subtask state, whereas those in Table~\ref{tab:ablation} are trained
without it. The comparisons hold only within the figure or within the
table.
The configuration shared by both (layers $3$--$20$, eight frames, the
full query set) reaches $49.3\%$ SR and $45.7\%$ SPL with the subtask
state and $42.3\%$ and $38.4\%$ without it, a gap of about seven points
on both metrics that we attribute to the subtask state.


\vspace{-1em}
\paragraph{History length and role queries.}
Table~\ref{tab:ablation} varies the number of history timepoints in the
spatial memory and the number of role queries.
SR first rises with the history length and then falls.
With the reduced query set, going from four to eight frames raises SR
by $3.5$ points and SPL by $3.6$ with the full set, going from six to
eight adds $2.0$ and $1.6$.
From four to eight frames NE and OSR barely change ($7.58$ to
$7.57$\,m, $+0.9$ points): the agent comes close to the goal about as
often but stops there more reliably, which suggests that a longer
history mainly helps the model judge its progress.
More frames are not always better, however with nine frames SR drops
back to $40.7\%$ and SPL to $36.9\%$.
The number of role queries matters as well. At eight frames, about four
times fewer queries lower SR by $2.0$  and SPL by $1.6$ points.
Unlike a shorter history, this also reduces how often the agent reaches
the goal: OSR falls by $2.6 \%$ and NE grows from $7.20$ to
$7.57$\,m.
The role queries carry what the decoder reads from the spatial memory to
the prediction heads, and we hypothesize that with too few of them, too
little spatial information is extracted through cross-attention to
localize the next waypoint.

\vspace{-1em}
\paragraph{Insertion depth.}
Fig.~\ref{fig:ablation_layers}(a) varies the decoder layers, out of
$36$, that read the memory.
Layers $3$--$20$ give $49.3\%$ SR and $45.7\%$ SPL.
Adding the first three layers ($0$--$20$) lowers SR by $6.1$ points, and
extending the range to layer $25$ lowers it by $4.7$. OSR and SPL follow
the same trend.
More layers are therefore not better: the smallest of the three sets
performs best.
We hypothesize that in the first layers the querying tokens are not yet
conditioned on the instruction, so the memory is read without task
context, whereas the last layers are closer to output prediction and
benefit less from new evidence.
\vspace{-1em}
\paragraph{Placement relative to self-attention.}
Within a layer, reading the memory before or after self-attention
performs similarly (Fig.~\ref{fig:ablation_layers}(b)): $49.3\%$ SR
against $48.7\%$, with OSR and SPL within one point.
We read before self-attention in all other experiments.

\begin{figure*}[t]
\centering
\includegraphics[width=1.0\textwidth]{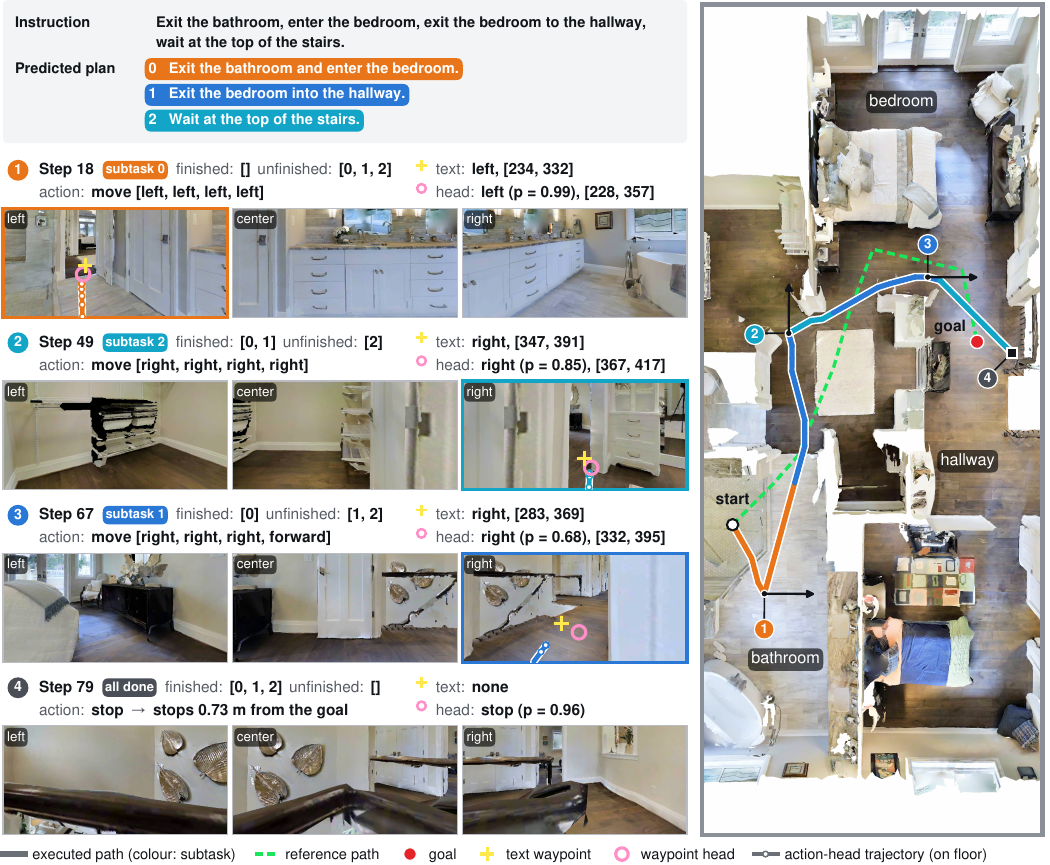}
\caption{\textbf{A successful episode in which the model corrects its subtask state.}
\emph{Left:} the instruction, the plan written by the model, and four
model calls. Each row prints the text answer and the Waypoint Head
output, and shows the three input views with the waypoint from the text
answer (cross), the waypoint from the Waypoint Head (ring), and the
Action Head trajectory projected onto the floor.
\emph{Right:} top-down view with the reference path and the executed
path, colored by the active subtask.
At step~49 the model has marked subtask~1 as finished too early (short
teal segment). By step~67 it has reopened it, and it stops $0.73$\,m
from the goal.}
\label{fig:qualitative}
\end{figure*}


\subsection{Qualitative Analysis}
\label{sec:qualitative}

Fig.~\ref{fig:qualitative} shows a successful R2R-CE val-unseen episode
in which the model corrects its own subtask state.
At its first call the model splits the instruction into three subtasks.
At step~49 the state lists subtasks~0 and~1 as finished, although the
agent has not yet reached the hallway that subtask~1 refers to.
By step~67 the state lists subtask~1 as unfinished again: the model has
reopened it after about nine steps ($1.0$\,m), which appears as the
short teal segment on the map.
The agent then leaves the bedroom, marks all three subtasks as finished,
and stops $0.73$\,m from the goal at step~79.
The carried state is therefore not simply copied forward. The model
revises it when later observations contradict it.
In the three calls that predict a waypoint, the text answer and the
Waypoint Head select the same view, and their pixels differ by $26$,
$33$, and $55$ in the $640\times480$ rendered frame.

\section{Discussion}
\label{sec:discussion}

\textbf{MaRSNav} keeps historical observations outside the VLM input sequence: a frozen VGGT-$\Omega$ encodes them, and the decoder retrieves spatial information through cross-attention, so the sequence does not grow with the history. Combined with explicit task-state tracking, it achieves $65.4\%$ and $66.5\%$ SR on R2R-CE and RxR-CE using only benchmark demonstrations for navigation fine-tuning. Our ablations show that cross-attention placement matters: reading the memory in decoder layers $3$--$20$ gives $49.3\%$ SR on R2R-CE, whereas also reading it in layers $0$--$2$ reduces SR by $6.1$ points.

Several limitations remain. First, the spatial memory covers only the last eight sampled timepoints ($28$ environment steps) and relies on a frozen geometry encoder without adaptive retention. Earlier observations are therefore no longer directly accessible, although their task-level implications may persist in the subtask state. Second, training uses only expert demonstrations, without policy-generated recovery trajectories, which may limit recovery from navigation errors and distribution shifts. On RxR-CE, \textbf{MaRSNav} trails OmniNav and CorrectNav ($66.5\%$ versus $73.6\%$ and $69.3\%$ SR), which use additional navigation data and recovery-oriented supervision. Finally, all evaluations are in simulation, leaving real-world generalization unverified.

In future work, an adaptive, episode-level memory could selectively retain and summarize observations by their relevance to the navigation context, giving access to spatial information beyond the fixed window. Recovery trajectories and more diverse navigation data could also improve robustness to off-trajectory states.
\section{Conclusion}
\label{sec:conclusion}

In this paper we introduced \textbf{MaRSNav}, a navigation grounding framework that combines context-aware spatial memory retrieval with persistent and revisable task-state tracking for continuous vision-and-language navigation. By constructing external spatial memory from multi-view observations using a pretrained geometry model, \textbf{MaRSNav} enables the VLM to retrieve historical spatial information through decoder-level cross-attention conditioned on navigation instructions, current observations, and task progress, without adding historical visual tokens to its input sequence. A structured subtask state maintains navigation progress across decisions, while state perturbation during training encourages the model to correct inaccurate progress estimates based on visual evidence. Furthermore, jointly supervised role queries coordinate image-space waypoint localisation and local trajectory prediction in a single forward pass. Experiments on R2R-CE and RxR-CE demonstrate competitive navigation performance in unseen environments using only benchmark demonstrations for navigation fine-tuning. 
\section*{LLM Usage Statement}
Qwen3.7-[Plus/Max] was used to produce the subtask annotations described in \cref{sec:subtask_state}: it split each training instruction into ordered subtasks and marked the trajectory frame at which each subtask is completed. GPT (OpenAI) and Claude (Anthropic) were used to assist with code development, polish the authors' drafts of the abstract, introduction, related work, and method sections, and support the design of figures and visualizations. Claude was also used to check the consistency of terminology and reported numbers. The authors designed the method, conducted all experiments, verified the manuscript content and reported results, and take full responsibility for the final submission.
{
    \small
    \bibliographystyle{ieeenat_fullname}
    \bibliography{main}
}


\end{document}